\documentclass[letterpaper, 10 pt, conference]{ieeeconf}
\IEEEoverridecommandlockouts     
\usepackage{times}
\usepackage{helvet}
\usepackage{courier}
\usepackage{color}
\usepackage{amsmath}
\usepackage{amssymb}
\usepackage[ruled,vlined,linesnumbered]{algorithm2e}
\usepackage{graphicx}
\usepackage{multirow}
\usepackage{graphics}
\usepackage{flushend}

\DeclareMathOperator*{\argmax}{argmax}

\newcommand{\commentAD}[1]{}

\SetKwFor{Loop}{Loop}{}{EndLoop}
\title{Waterberry Farms: A Novel Benchmark For Informative Path Planning}
\author{Samuel Matloob, Partha P. Datta, O. Patrick Kreidl, Ayan Dutta, Swapnoneel Roy and Ladislau B{\"o}l{\"o}ni
\thanks{Partha Datta, O.P. Kreidl, A. Dutta, and S. Roy are with the University of North Florida, USA. Emails: {\tt\small \{n01490529, patrick.kreidl, a.dutta, s.roy\}@unf.edu}
\newline 
S. Matloob and L. B{\"o}l{\"o}ni is with the University of Central Florida, USA. Emails: {\tt\small sam.matloob@knights.ucf.edu, ladislau.boloni@ucf.edu}\newline
This work is supported in part by NSF CPS Grants \#1932300 and \#1931767.
}%
}

\begin{document}
\maketitle

\thispagestyle{empty}
\pagestyle{empty}

\begin{abstract}
Recent developments in robotic and sensor hardware make data collection with mobile robots (ground or aerial) feasible and affordable to a wide population of users. The newly emergent applications, such as precision agriculture, weather damage assessment, or personal home security often do not satisfy the simplifying assumptions made by previous research: the explored areas have complex shapes and obstacles, multiple phenomena need to be sensed and estimated simultaneously and the measured quantities might change during observations. The future progress of path planning and estimation algorithms requires a new generation of benchmarks that provide representative environments and scoring methods that capture the demands of these applications. 

This paper describes the Waterberry Farms benchmark (WBF) that models a precision agriculture application at a Florida farm growing multiple crop types. The benchmark captures the dynamic nature of the spread of plant diseases and variations of soil humidity while the scoring system measures the performance of a given combination of a movement policy and an information model estimator. By benchmarking several examples of representative path planning and estimator algorithms, we demonstrate WBF's ability to provide insight into their properties and quantify future progress. 

\end{abstract}


\section{Introduction}

The increased availability and affordability of unmanned aerial vehicles made possible their deployment in information collection applications such as investigating structural damage in the roofs, surveying invasive species, tracking pollution, responding to natural disasters, and precision agriculture. Finding a path that enables a set of favorable observation positions is the goal of {\em informative path planning} (IPP)~\cite{hollinger2014sampling,nguyen2014nonmyopic,singh2009efficient,wei2020informative}. Multi-agent informative path planning (MIPP) extends the problem to multiple drones, subject to various communication constraints~\cite{dutta2021multi,Samman2021Secure,kantaros2021sampling,ma2018multi,viseras2018distributed}. 

IPP has a rich literature ranging from theoretical analysis to engineering case studies~\cite{cao2013multi,krause2008near,singh2009efficient}. A wide range of sophisticated path planning algorithms were developed over the course of several decades. Most real-world deployments, however, use the simplest possible models such as lawnmower or spiral motion patterns. There are several reasons for the lack of adoption of more complex IPP models, but we contend that the primary reason is the difficulty to assess how much benefit would a complex, state-of-the-art algorithm give for a given problem. All variations of IPP are inherently difficult -- even under highly simplifying assumptions, they reduce no further than the already NP-complete vertex cover problem~\cite{arkin2000approximation}. Unfortunately, in many practical deployments, the real-world conditions force significant departures from the simplifying assumptions. Adding realistic models of observation, environment, information, communication, and movement adds not only further complexity but also changes the relative rankings of algorithms. Finally, the scale required for practical applications limits the type of algorithms that can be realistically considered as solutions. 

There is thus a need for benchmarks that capture the requirements of the new generations of applications, can validate the performance of different algorithms, and catalyze future IPP research. The Waterberry Farms (WBF) suite of benchmarks is based on a realistic formulation of the IPP problem inspired by the needs of precision agriculture. We assume a farmer on a typical-size Florida farm, who is growing tomatoes and strawberries. The farmer is using one or more drones to inspect the farm, detect and track the spread of plant diseases, and evaluate soil humidity. The typical size of a Florida farm and the necessity of close inspection for the detection of plant diseases naturally determine the scale of the problem. The fact that plant diseases spread and the soil humidity changes over time, requires us to consider the temporal dimension in sensing when the observations happen over multiple days. The mixed cultivation patterns, the relative importance of different diseases, and the typical geometric layouts of farms add additional complexity.

The benchmarks, together with their supporting code, environment, and information models, scoring functions, sample path planning, and estimation functions are released under the open source MIT license at 
{\tt \small https://github.com/lboloni/MREM}.

The contributions of this paper can be summarized as follows. We describe and analyze a precision agriculture application scenario and formalize it as an IPP challenge problem. We describe the WBF benchmark suite, its implementation, simplified and scaled-down versions, and scoring. Finally, we perform a series of experiments by running several representative path planning and estimation algorithms on the WBF benchmark. These experiments show that there is a significant difference in the scores achieved by different path planning and estimation techniques. We also find that Gaussian Processes~\cite{rasmussen2003gaussian}, the most popular estimation technique in the IPP literature~\cite{cao2013multi,hitz2014fully,krause2007nonmyopic,singh2009nonmyopic,ma2017informative,viseras2017online,wei2020informative}, outperforms naive estimation techniques in static settings and small scenarios, but in its canonical form does not scale to IPP problems of practically useful size.

\section{Related Work}

Finding a solution to variations of the IPP problem attracted a significant research effort over the last decades. Chekuri and Pál~\cite{Chekuri-2005-Greedy} proposed a greedy algorithm that finds a route that maximizes rewards in a graph problem.  
Singh et al.~\cite{Singh-2007-Efficient} proposed a variant of Chekuri and Pál's algorithm adapted to the MIPP setup, using a spatial decomposition model. Karaman et al.~\cite{Karaman-2011-ICRA} proposed a single robot path planning, extending the Rapidly-exploring Random Tree (RRT*) algorithm. Hollinger and Sukhatme~\cite{hollinger2014sampling} developed an algorithm that extends Rapidly-exploring Random Graphs (RRGs) with a branch and bound technique to find the optimal path and prune paths that exceed the requirement. Cao et al.~\cite{cao2013multi} describes two approaches based on entropy and mutual information criteria for the MIPP problem. The paper assumes that the environmental information is affected by an anisotropic field phenomenon that gives more spatial correlation measurements in a specific direction. The anisotropic field model can represent natural phenomena such as plankton density and fish abundance in the oceans, temperature, salinity, peat thickness, etc. In this work, the robot's path relies on the transect sampling, which follows the direction that exhibits a higher spatial correlation. 

Hitz et al.~\cite{hitz2014fully} proposed LSE-DP (Level Set Estimation - using Dynamic Programming) to classify measurements and determine an efficient path for a robot that reduces uncertainty around areas within the threshold value. The proposed approach was applied to monitoring the level of cyanobacteria in a lake using an Autonomous Surface Vessel (ASV). Ma et al.~\cite{ma2018multi} considered the MIPP problem in the context of multiple objectives such as travel cost minimization, information gain, and multi-robot coordination. The proposed system allows the robots to update their path function of the newly collected data. The sample application considered was the use of autonomous underwater vehicles (AUVs) to monitor the ocean's temperature, salinity, and chlorophyll contents. 

Kemna et al.~\cite{kemna2017multi} used Voronoi partitioning to lay out the boundary for each robot in a multi-AUV setting, with the partitions periodically updated as more information is available. Luo et al.~\cite{luo2018adaptive} also considered adaptive sampling for multi-robot sensor coverage using a mixture of Gaussian Processes. Singh et al. \cite{singh2009nonmyopic} proposed a compromise between exploration and exploitation in adaptive IPP. The approach in their algorithm was nonmyopic (which means that the algorithm will plan for possible and future observations) and exploited the sub-modularity (i.e., diminishing return over time) property of the objective function. The work was extended to the MIPP setting using the sequence-allocation algorithm.

\section{Formalization}
\label{sec:Formalization}

Informally, the IPP problem can be stated as finding the path of a robot such that the collected useful information is maximized, with the MIPP problem extending this to the case of multiple robots. 

In practice, MIPP formulations range from an abstract graph model to a physical robot control problem. The formalism we choose for the WBF benchmark identifies as the ultimate goal of MIPP -- build an information model that supports the decision-making of the customer. An immediate consequence of this view is that the path planning process must be always benchmarked in tandem with the estimation process used to build and update the information model. Naturally, one can always compare different path planners with a fixed estimator (or, different estimators with a fixed planner), allowing most existing algorithms to be conveniently plugged into the model. Nevertheless, pairs of estimators and path planners that are aware of each other's properties might have a non-trivial advantage. 

Starting from these considerations, we formalize the MIPP problem as follows. Let us consider a geographical area of size $x_\textit{max} \times y_\textit{max}$ where every location is denoted by its integer coordinates $(x,y)$ and a discrete temporal dimension $t$. At every location, at every timepoint, we can measure $k$ distinct quantities $m_i~~ i \in 1 \ldots k$, where the values are real numbers denoted with $E(x, y, t, m) \in \mathbb{R}$. We call the totality of measurable values encapsulated by the four-dimensional tensor $E$ the {\em environment}. We will use the Python \textit{numpy}-style ":" indexing notation to refer to slices of the tensor. Thus, $E(:, :, t, m)$ represents the scalar field of the measurement $m$ at time $t$, while $E(x, y, :, m_j)$ is the time series of the measurements for $m_j$ at location $(x,y)$. 

We consider that there are $n$ robots $R_1 \ldots R_n$ that move in the environment, with $pos(R_i, t) \rightarrow (x_{ri},y_{ri})$ being the location of robot at a given moment in time. At each timestep $t$, a robot can take multiple {\em actions}. In the version of WBF we describe here, we assume that each robot takes two actions: an {\em observation action} that creates an observation $o = E(x_{ri},y_{ri},t,:)$, and a {\em movement action} that determines the location of the robot at time $t+1$. The extension of this formalism to other kinds of actions (such as communication) and observation models (such as noisy observations, partial sensor suites, and extended observation ranges) is immediate, but beyond the scope of this paper. 

Thus, the operation of the $n$ robots in the environment for a timespan of $t$ will yield a collection of $n \times t$ observations $O = {o_{1,1}, \ldots o_{t,n}}$. The observations are used by an estimator $\theta(O) \rightarrow I$ to create an {\em information model} $I$ that has the same structure as the environment $E$. In general, we want an information model that is a good approximation of the environment according to a loss function $\mathcal{L}(E,I)$. We will call the negative loss function $-\mathcal{L}$ the {\em score}, and following the usual usage in machine learning, we will refer to maximizing the score. 

With these preliminaries, we are ready to define the MIPP problem. Let us consider a collection of environments $E$ sampled from a distribution $P(E)$. We call the {\em optimal pair of an estimator and a movement policy} the pair $\theta^*, \pi^*$ that maximizes the expectation of the score over the distribution of the environments considered:

\begin{equation}
    \theta^*, \pi^* = \argmax_{\theta, \pi} \mathop\mathbb{E}_{E \sim P(E)} - \mathcal{L}(E,I(\theta,\pi))
\end{equation}

The distribution $P(E)$ captures both the probabilistic dynamics of the environment quantities, as well as their spatial dependence. It is reasonable to assume that the estimator and movement policy has access to models of this distribution, or at least to samples that can be used as training data. 

\section{Design of the Benchmark}

\begin{figure}
\centering
\includegraphics[width=\columnwidth]{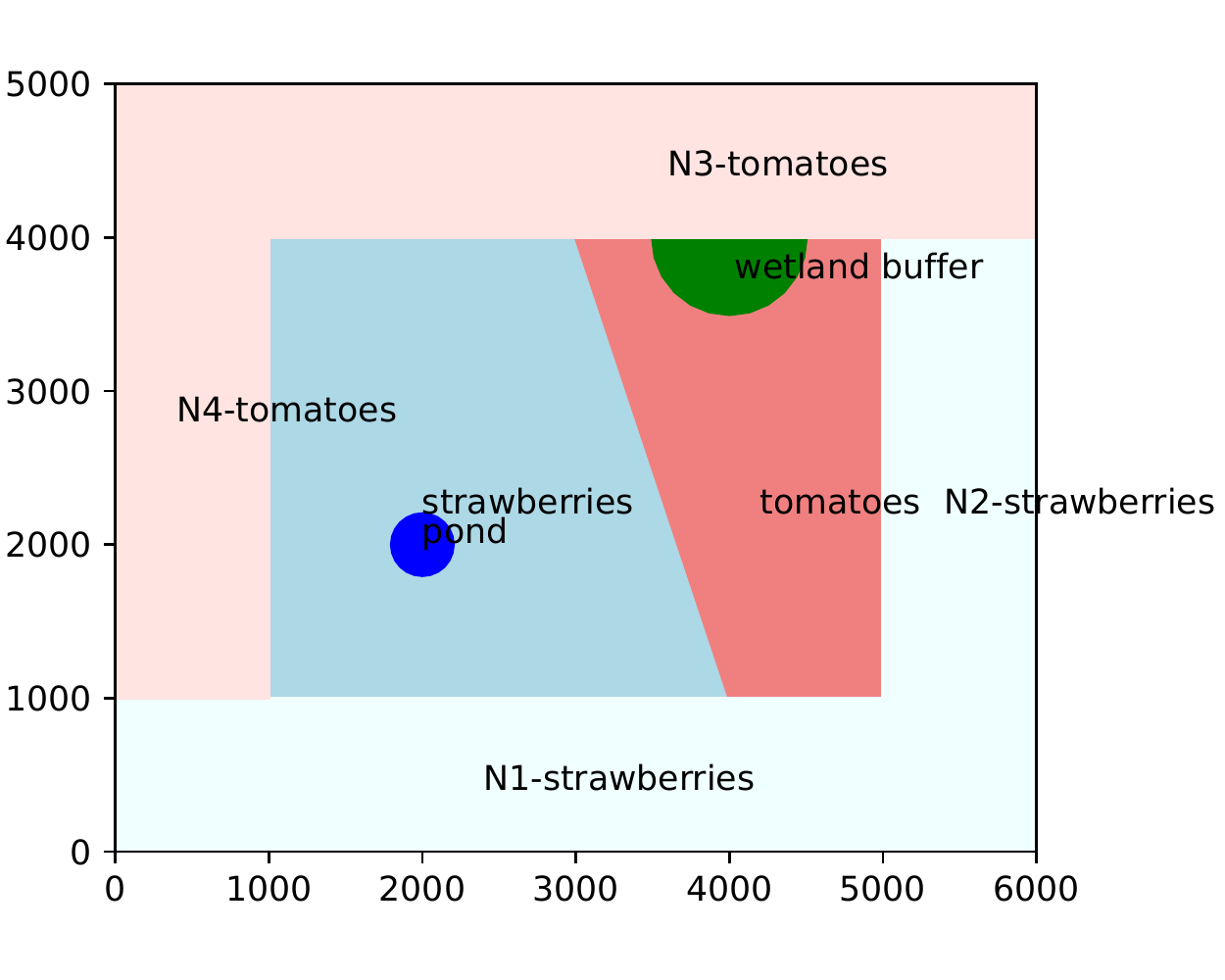} \\
\includegraphics[width=0.7\columnwidth]{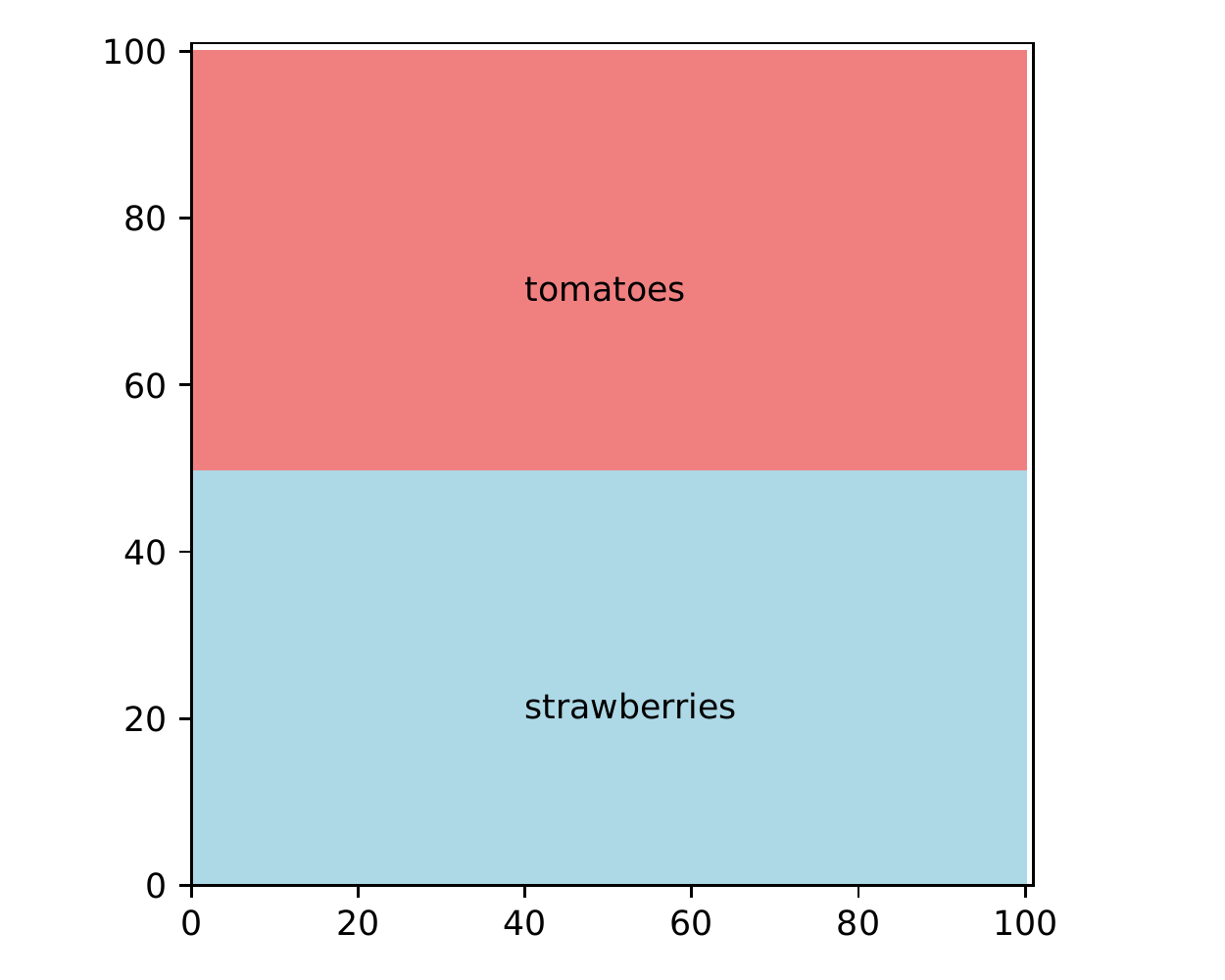}
\caption{
(top) The layout of Waterberry Farm and its neighbors. The farm extends between coordinates 1000 to 5000 on the x axis, and 1000 to 4000 on the y axis. Light blue areas of various shades are strawberry crops, while coral-colored areas are tomato crops. The areas marked N1...N4 are neighboring agricultural areas. (bottom)
The layout of the simplified Miniberry-100 geometry.
}
\label{fig:Geometry}
\end{figure}

The Waterberry Farms environment models a typical farmstead in the Central Florida area. It was designed as a composite of the layout, size, and land features of several farms that were on sale in January 2022 on the website Landwatch.com which specializes in the sale of rural properties. The overall layout of the farm and its immediate neighbors is shown in Figure~\ref{fig:Geometry}-top. The geographic features include areas suitable for agriculture, a pond, and a wetland buffer area (not suitable for agricultural cultivation). The farmer plants two of the most popular crops in the area: strawberries and tomatoes, with the crops separated by a natural separation line -- a publicly accessible road. The farm is adjacent to other farms with their own crops of strawberries and/or tomatoes. 

For the benchmark, we consider three quantities of interest to the farmer: soil humidity (in all cultivated areas), the presence of tomato yellow leaf curl virus TYLCV (on the tomato crops)~\cite{moriones2000tomato}, and the presence of charcoal rot CCR (on the strawberry crops)~\cite{mertely2005first}. The early detection of TYLCV requires the detection of the moment when several leaves of the plant turned yellow and acquired a curl. The spread of TYLCV can lead to a total loss of the tomato crop. CCR can be recognized by the older leaves of the strawberry plant turning gray-green in color and beginning to dry up. CCR progresses slower than TYLCV and the loss of crop is only partial. As both diseases require the close-up inspection of individual leaves, we will consider one square foot as the individual unit of observation and robot movement. Thus, the WBF environment can be modeled in our formalism by taking $x_\textit{max}=6000$, $y_\textit{max}=5000$ and $m=3$. While the farmer is only interested in the information from its own land, the consideration of the neighbors is necessary for the modeling as crop diseases spread across ownership boundaries. 

While the goal of the WBF benchmark is to offer a way to benchmark path planning and estimator algorithms in a realistic setting, we also need to offer an on-ramp
for the measurement and improvement of current algorithms. The Waterberry geometry is unrealistically difficult for many current estimators and path planning algorithms. For instance, with 5000 x 6000 x 3 = 90M datapoints in the model, implementations of the original Gaussian Process algorithm with its $O(n^3)$ inference complexity cannot be used as an estimator. This problem is also unusually large for the current generation of deep RL techniques, which are often benchmarked on the Arcade Learning Environment~\cite{bellemare2013arcade} with its 210x160 setting. To facilitate the scaling up of these techniques, we designed several variations of the benchmark with reduced difficulty. Three simplified maps, called Miniberry-10, -30 and -100 respectively reduce the size of the map to squares of the specified size, while also eliminating the pond, wetland, and the neighbor models (see Fig.~\ref{fig:Geometry}-bottom).

To capture the possible distributions of the environments $P(E)$ we must rely on domain-specific knowledge about the natural dynamics of the modeled phenomena. TYLCV and CCR are infectious plant diseases that spread, with certain probabilities, to nearby plants. 

The epidemic spread model extends the standard SIRV model of epidemics~\cite{shulgin1998pulse,kermack1927contribution} to a 2D physical environment of an agricultural field. Each 1x1 ft cell of the agricultural area can be in one of the S, I, R or V states. An S cell is susceptible - this is normally a healthy plant. The probability for a susceptible cell to become infected depends on the cells in the neighborhood with further cells having a lesser influence. We assume that the infection probabilities can be created with a convolution based on a distance matrix. Each individual plant follows a SIR cycle. An infected cell I remains infectious for a specific amount of time, which is a parameter of the model. An R cell is a cell that contains plants that are dead and thus cannot spread the disease further. A vaccinated cell V in the original model is a cell that is not susceptible. In the context of the WBF model, the V cells correspond to areas that are not planted with a plant that is susceptible to the disease. For instance, for the TYLCV, the areas planted with strawberries, as well as ponds and wetlands count as V-type.

As implemented, the model can be parameterized by the total likelihood of propagation and by the size and content of the propagation probability kernel. For TYLCV and CCR these values were adjusted to achieve a model that intuitively corresponds to the typically observed spread of these diseases. 

As these diseases are plant-specific, the epidemic spread model is only applied through the areas where a specific crop is grown (tomato for TYLCV and strawberry for CCR), and they do not spread over the pond or wetland areas. However, the disease spreads across the boundaries of the property, thus the disease might spread from the neighbor's property to the farmer as long as they grow the same crop. 

The soil humidity model is based on a precipitation/evaporation model. It assumes that the soil loses humidity at a constant rate through evaporation, and acquires humidity through periodic showers that affect certain areas. We model these areas and the precipitation through a mixture of Gaussians with means $\mu = [\mu_x, \mu_y]$ distributed uniformly in the area, and a covariance matrix $[ (h/8)^2, 0 ; 0, (w/8)^2]$ where $h$ and $w$ are the effective sizes of the area affected by the shower. \commentAD{how is the evaporation part modeled?}

\begin{figure}
\centering
{\footnotesize TYLCV Miniberry-100}\\
\includegraphics[width=\columnwidth]{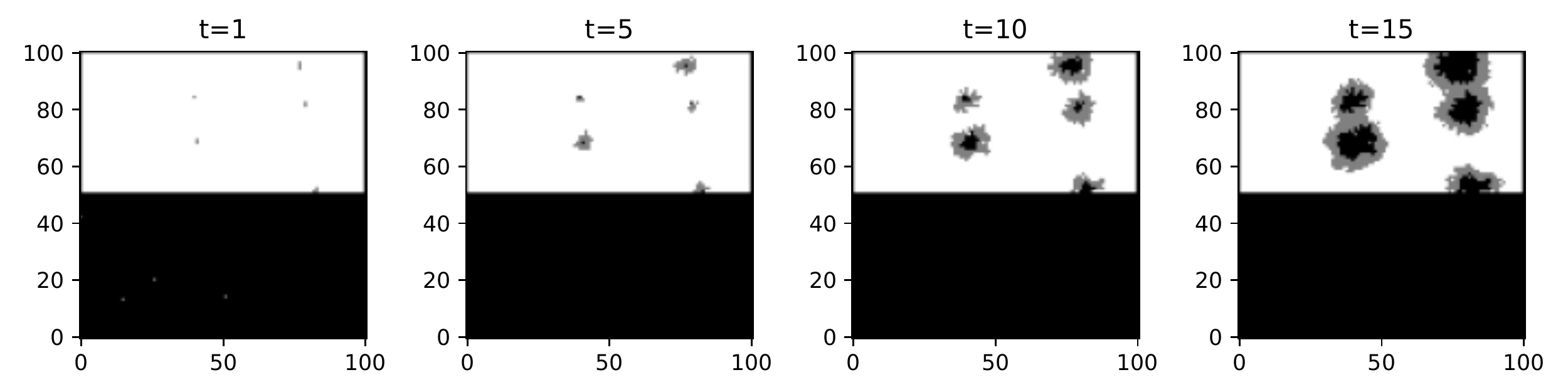}\\
{\footnotesize CCR Miniberry-100}\\
\includegraphics[width=\columnwidth]{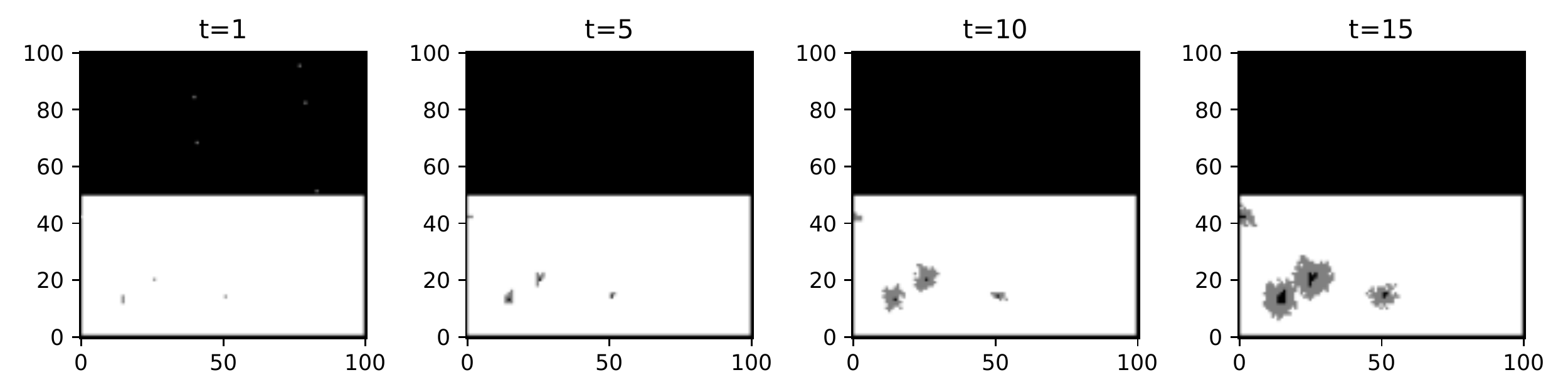}\\
{\footnotesize Soil humidity Miniberry-100}\\
\includegraphics[width=\columnwidth]{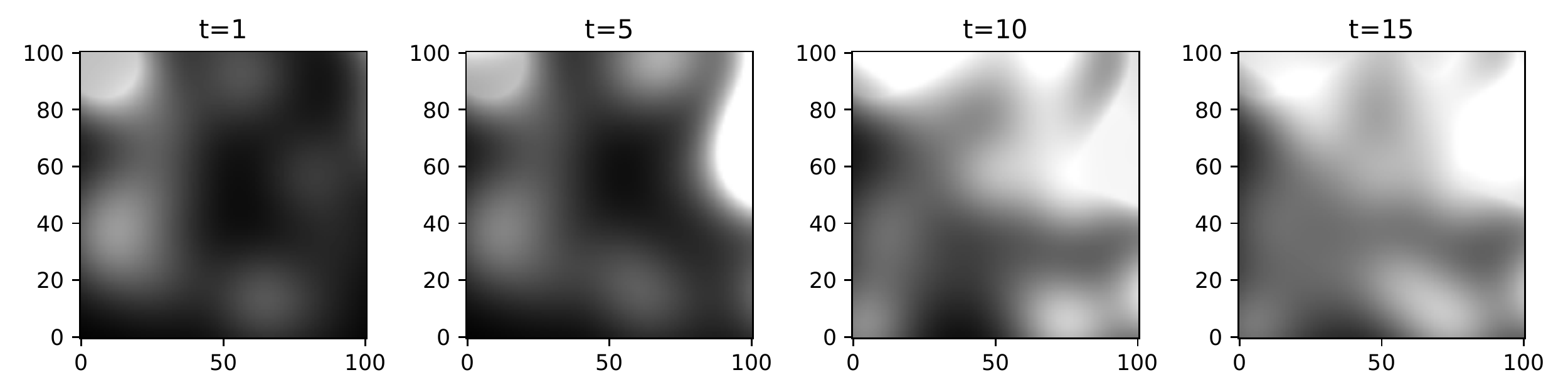}\\
{\footnotesize TYLCV Waterberry}\\
\includegraphics[width=\columnwidth]{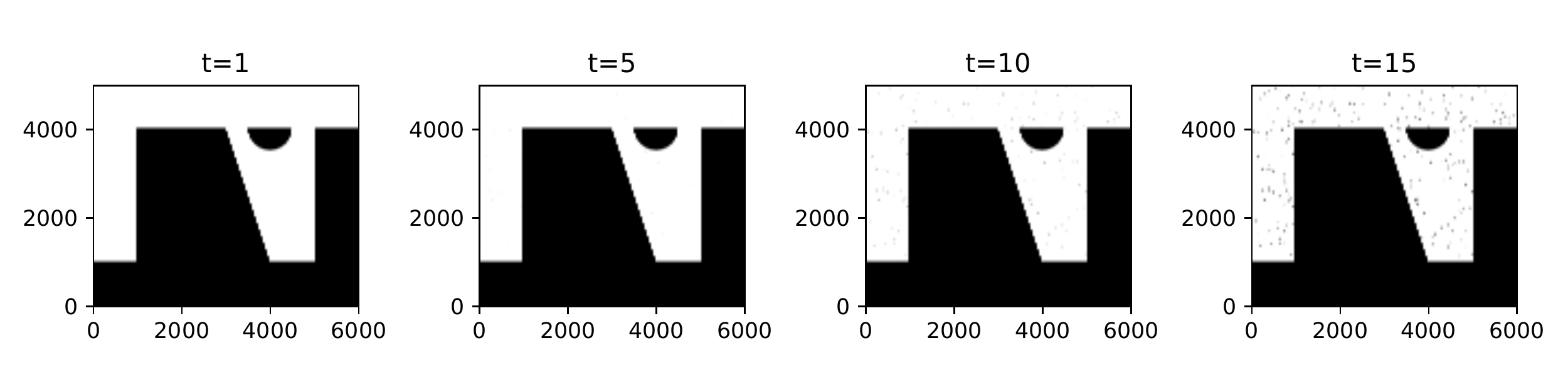}\\
{\footnotesize CCR Waterberry}\\
\includegraphics[width=\columnwidth]{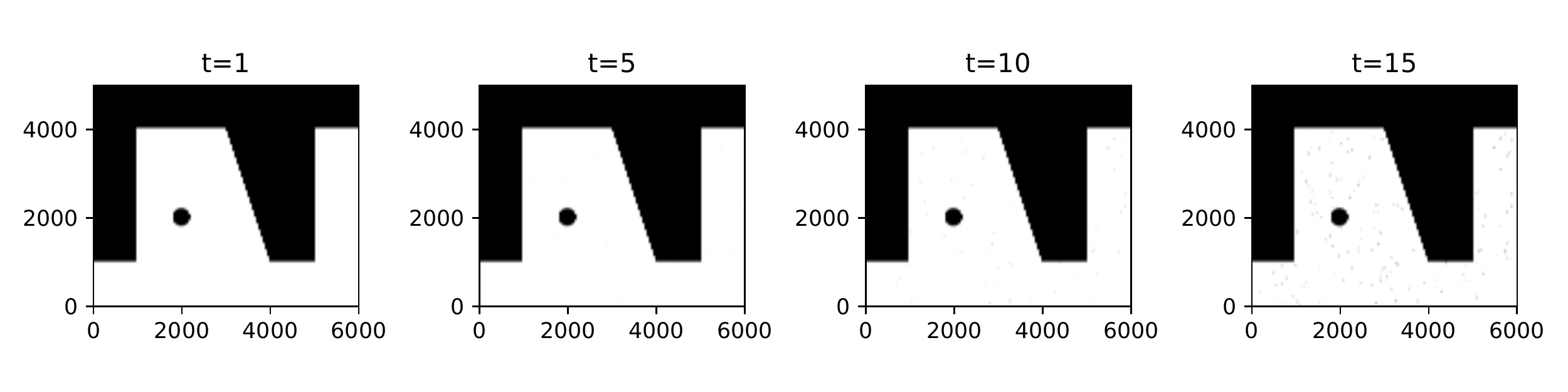}\\
\caption{Examples of the temporal dynamics of the TYLCV, CCR and soil humidity metrics for the Miniberry-100 and Waterberry environments. The black areas on the TYLCV and CCR models represent areas that are not susceptible to the diseases. }
\label{fig:Dynamics}
\end{figure}

Fig.~\ref{fig:Dynamics} shows examples for the dynamics of the environments over the course of a 15-day span. 

\subsection{Observations and scoring}

In the WBF benchmark, at every timestep, the robot(s) make three observations concerning the TYLCV and CCR diseases, and soil humidity. Based on this information, the robot must update a composite information model that estimates these values for each part of the map. 

The scoring of the benchmark is exclusively based on the {\em accuracy} of the estimates; the {\em quantity} of the collected information affects the score only indirectly. The maps are masked for relevance: we are only interested in the TYLCV for areas planted with tomatoes and CCR for areas planted with strawberries, while we are interested in soil humidity everywhere except the unplanted areas in the Waterberry model. The error is weighted based on the economic value of the estimate: information about the TYLCV disease, which can lead to a total loss of the tomato crop is weighted higher than CCR. For TYLCV and CCR we are using an asymmetric model with a higher cost for the false negatives compared to the false positives. The soil humidity value has a symmetric loss model.

The overall score model can be described with the following equation: 

\begin{IEEEeqnarray*}{l}
\mathcal{L}(E,I) = \\
\frac{\sum\limits_{i=1}^{k} w_i \sum\limits_{(x,y)} M(x,y,i) \sum\limits_{l=1}^{c}  \textit{AE}(E(x,y,t_{Ll},i), I(x,y,t_{Ll},i))}{c\cdot \sum\limits_{i=0}^{k}w_i \sum\limits_{x=1}^{x_\textit{max}} \sum\limits_{y=1}^{y_\textit{max}} M(x,y,i)} 
\end{IEEEeqnarray*}

Here the $w_i$ is the overall importance weight for the measurement $i$. $M(x,y,i)$ is the relevance mask for the measurement $i$, having a value of 1 if the measurement is relevant (eg. tomato is planted) and 0 if the measurement is not relevant (e.g. no tomato is planted, or the area not owned by the client). The asymmetric error AE is described by the formula:
\begin{equation*}
AE(a,b) =
\begin{cases}
c_{+} \cdot (a-b)^2 & \text{if } a < b,\\
c_{-} \cdot (a-b)^2 & \text{otherwise} 
\end{cases}
\end{equation*}

\noindent where $c_+$ and $c_-$ are the weights of negative and positive errors respectively. The following table shows the parametrization used by the WBF implementation: 

\bigskip

{
\small
\begin{tabular}{llccc}
    Name & Type & w & $c_-$ & $c_+$ \\
    \hline 
     TYLCV & Asymmetric & 1.0 & 1 & 10 \\
     CCR & Asymmetric & 0.2 & 1 & 10 \\
     Soil humidity & Symmetric & 0.1 & 1 & 1 \\
\end{tabular}
}

\bigskip

The mask values $M(x,y,i)$ are extracted from the geometries of the various models as shown in Fig.~\ref{fig:Geometry}, and they correspond to the areas in Fig~\ref{fig:Dynamics} that are not masked in black. The score masks do not consider areas owned by neighbors that are planted with tomatoes and strawberries, while the propagation models do consider these.
\section{Experiments}
\label{sec:Experiments}

In the following experimental studies, we use the WBF benchmark to study the impact of the planning algorithms and estimators on the performance of information collection as well as the computational cost of various estimators and its impact on the performance. Our objective is to investigate whether different path planner / estimator pairs perform at a sufficiently different performance to justify future research in these algorithms as well as extensive benchmarking. All the implementations are done in Python relying on the standard numpy, scipy and scikit-learn libraries. 

\subsection{The performance of path planner / estimator combinations}

\begin{figure*}
\includegraphics[width=\textwidth]{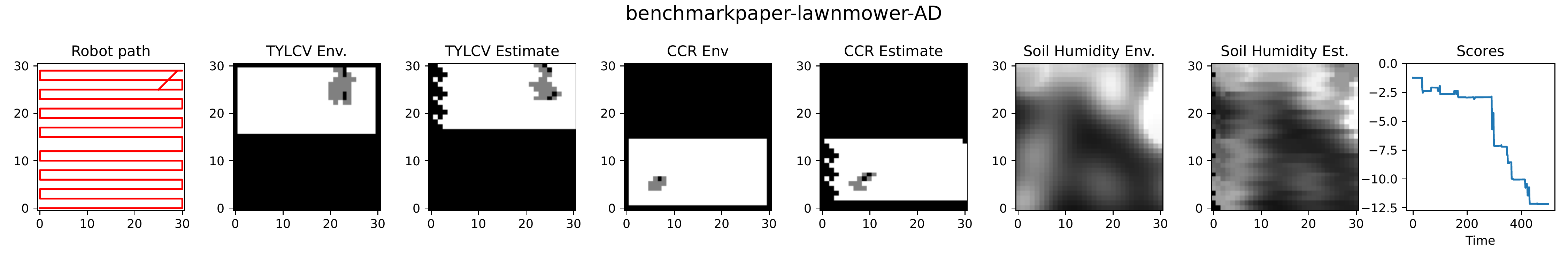}
\includegraphics[width=\textwidth]{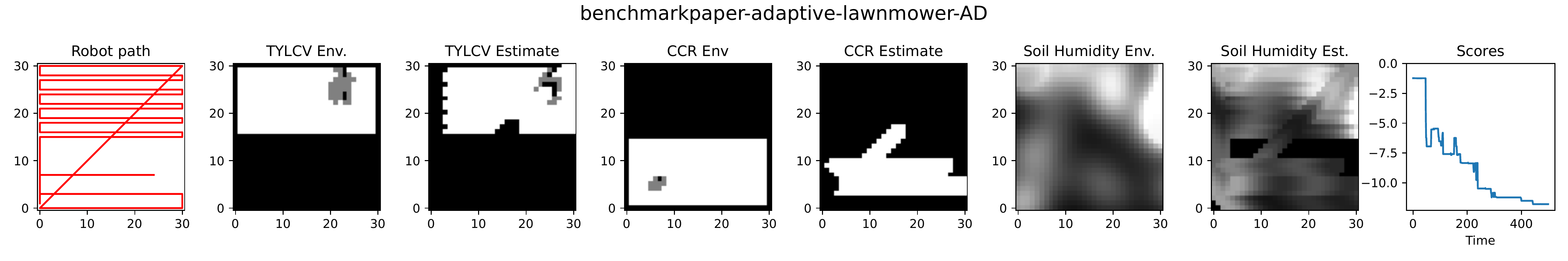}
\includegraphics[width=\textwidth]{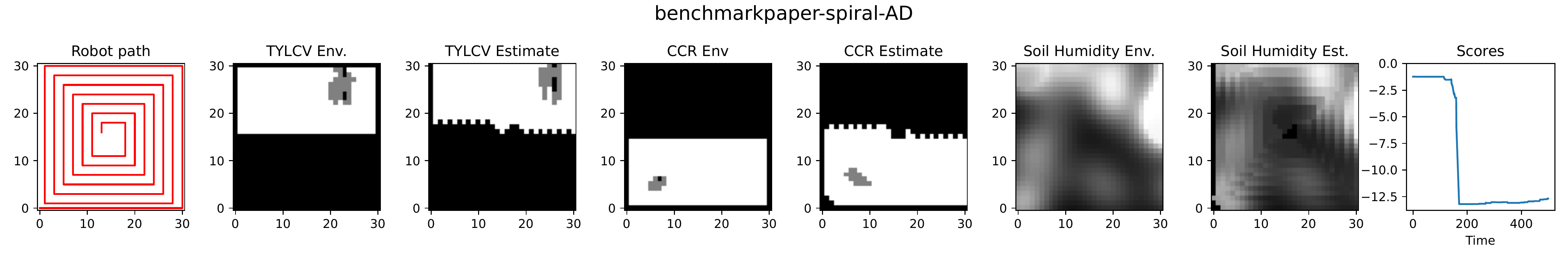}
\includegraphics[width=\textwidth]{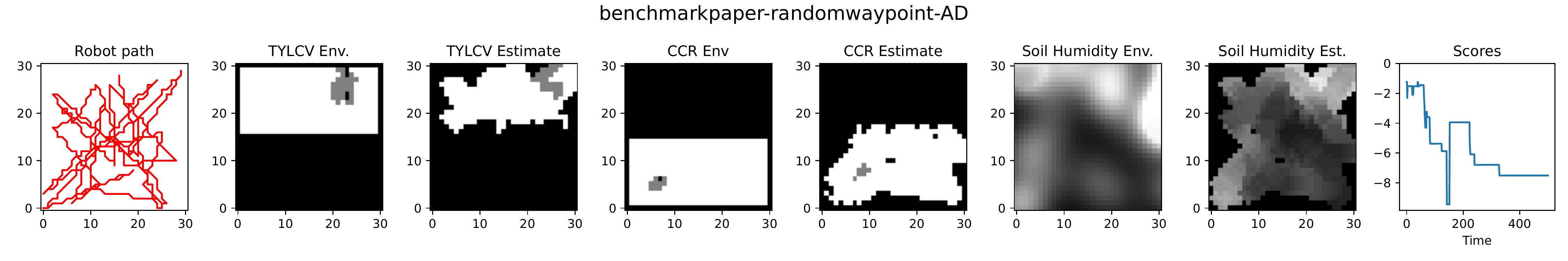}
\includegraphics[width=\textwidth]{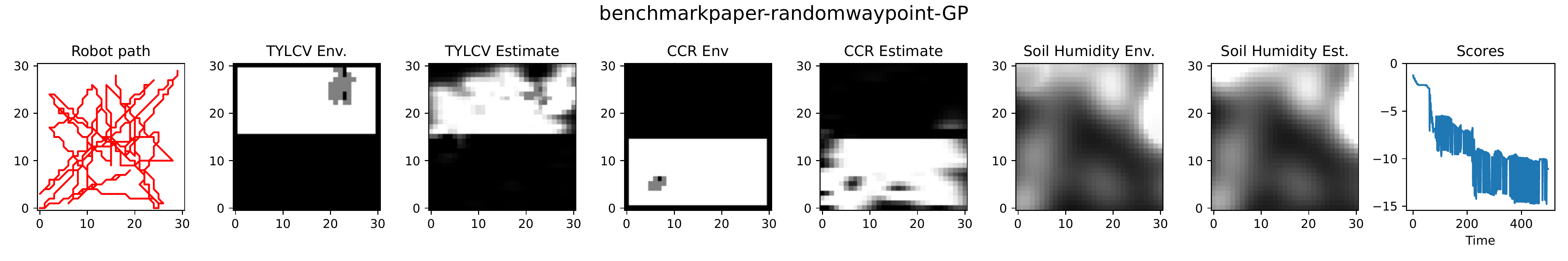}
\caption{Benchmark results for Miniberry-30, for a single day scenario. For each row, we show the path of the robot, the environment and information model values, and the evolution of the score in time. }
\label{fig:Comparison}
\end{figure*}

In this series of experiments, we run the Miniberry-30 benchmark using combinations of four path planning algorithms and two estimator models. We considered a scenario length of 500 timesteps. While very generous for the 900 grid cells of the Miniberry-30 model, this is motivated by the fact that we are calculating and plotting the score for every timestep, and the long scenario allows us to better study the evolution of the performance of the planner/estimator pairs.  

We benchmarked four path planning algorithms. The {\tt lawnmower} algorithm aims to cover the full area with a lawnmower-type back-and-forth pattern during the mission. The {\tt adaptive-lawnmower} path-planner adapts its exploration to the crop and the relative value of the information. Thus, it first covers and allocates more time to the tomato crop where the turns of the lawnmower will be denser. The {\tt spiral} path planner explores the area in a rectangular spiral pattern, with the density of the spiral being calculated such that the spiral is completed in the available time. Finally, the {\tt random-waypoint} approach is based on choosing a waypoint in a uniform random manner in the field, moving to it, and repeating the process until the time runs out. 

We combined the four path planners with a relatively simple estimator we call Adaptive Disk (AD). This assumes that an observation is valid in a disk of radius $r$, with the value of $r$ decreasing during the experiment as more observations are added. We also compared the performance of the {\tt random-waypoint} algorithm with a Gaussian Process (GP)-based estimator. The GP-based estimator uses the implementation in the scikit-learn library. After experimenting with various parameter values, we have used a kernel that is a combination of the radial basis and white noise kernels; and we have allowed a maximum 5 restarts for the optimizer. 

The results for the five combinations of path planning and estimator algorithms are shown in Fig.~\ref{fig:Comparison}, and allow us to draw several interesting conclusions. First, the four approaches using the AD estimator all converge to a score of about -12 at the end of the 500 estimation steps, but the evolution of the score varies widely. In general, the score decreases (that is, improves) with time, but anomalies when the score gets worse, appear both for AD and GP. The algorithm which performs best in the early part of the scenario is {\tt adaptive-lawnmower}, an informed algorithm that aimed to collect the most valuable information first. The {\tt random-waypoint} algorithm also improved the score much faster in the early part of the exploration compared to the ``systematic'' algorithms, but ultimately ended up with a worse score due to missed areas in the exploration. Between the two systematic algorithms, the {\tt spiral} model improved the score faster than the {\tt lawnmower}, but both of them converged to similar final values. 

The GP estimator coupled with the random waypoint algorithm reached a better score eventually (at approximately -15), while presenting a significant variation from observation to observation.

\subsection{Computational cost of the estimator}

The last row of Fig.~\ref{fig:Comparison} shows that a better estimator, such as GP, can significantly contribute to the overall score achieved
However, such an estimate also comes with a significant computational cost. While the experiments in Fig.~\ref{fig:Comparison} were done on a scaled down model (Miniberry-30), the benchmark run for the random waypoint / GP combination took 43 minutes, in contrast to several seconds for the other choices. Clearly, the estimation cost can be a significant component in the choice of algorithms suitable for practical applications. 

We need to consider both the computational complexity of certain algorithms in the theoretical sense, as well as the practical wall-clock time of running them. It is known that the original GP algorithm has a computational complexity of $O(n^3)$, while the complexity of our AD implementation is $O(n)$. Note, however, that there is also a significant overhead for the querying of the model for every point of the environment. However, beyond this, the overhead components and multipliers can also affect whether the algorithm is suitable for practical deployment. 

\begin{figure}
\includegraphics[width=\columnwidth]{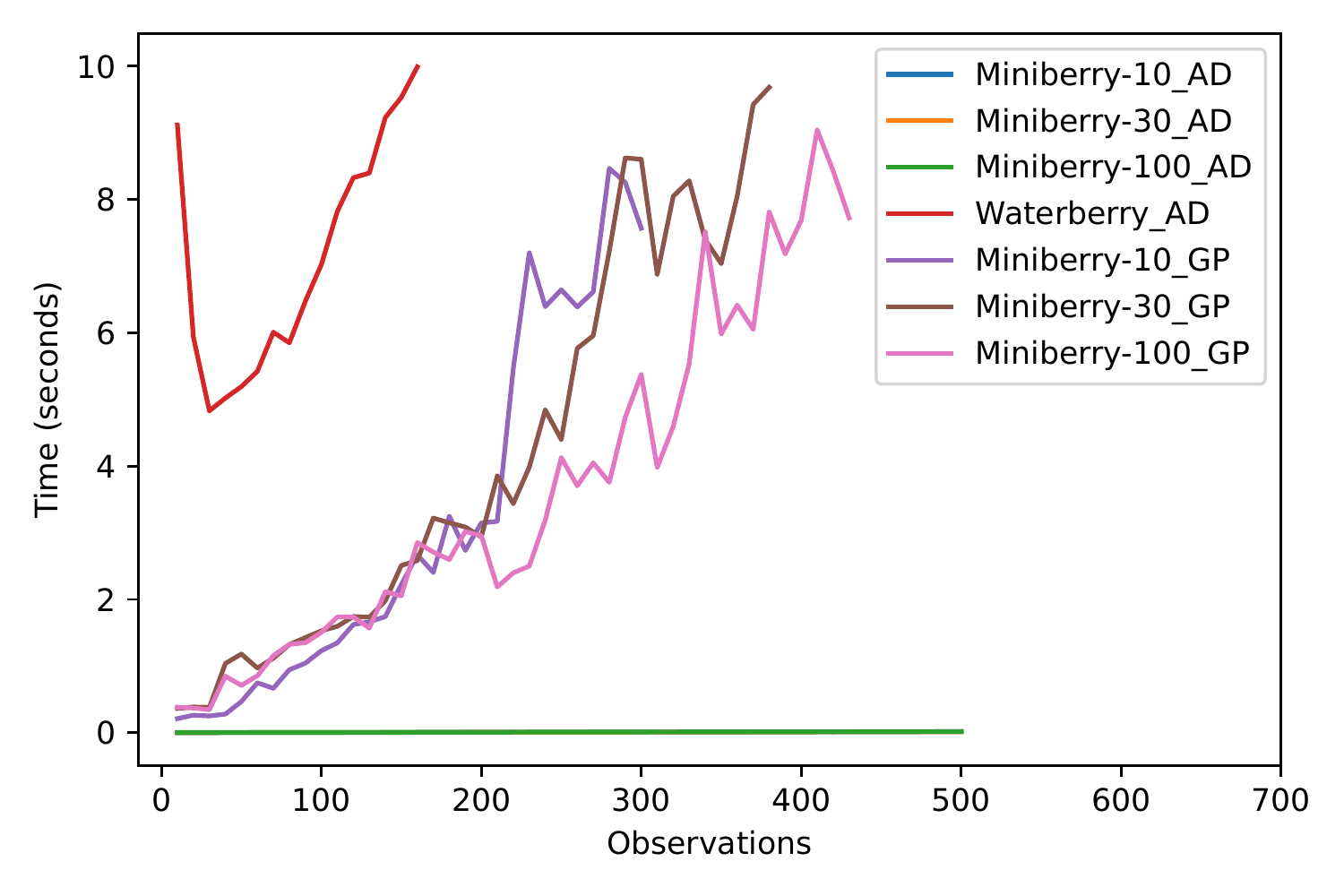}
\caption{Wall-clock time for the AD and GP estimators. The measured time includes the creation of three components of the information model for various geometries.  }
\label{fig:ComputationalCost}
\end{figure}

Fig.~\ref{fig:ComputationalCost} measured the wall clock time for a single estimation run for different estimators and geometries. For the sake of a cleaner representation, we stopped the experiments when one estimation run exceeded 10 seconds, which for this experiment, we define as the ``feasible'' limit. The experiments were done on a workstation with a 32-core AMD Threadripper processor and 64GB of memory. 

We find that the computation cost is negligible for the AD estimator on Miniberry 10, 30, and 100, and it is feasible on the full Waterberry benchmark (with 90M data points) for up to about 200 observations. For the GP algorithm, however, the computational cost is significant, and it escalates quickly. We had seen that the cost is already significant for Miniberry-30. With Miniberry-100, the costs remain feasible until about 450 observations - this, however, represents a very sparse coverage, such as a lawnmower pattern with only two turns. In the full Waterberry scenario, the GP algorithm does not return a solution in a feasible time even for a single observation. 

\section{Conclusions}

In this paper, we describe the Waterberry Farms benchmark, a framework to measure the performance of solutions to the IPP/MIPP problems, in the form of a movement policy and an environment estimator pair. Our experiments show that the benchmark allows us to gain unexpected insights into the performance and temporal behavior of various algorithms.  Among the insights with more general applicability that we gained is that the performance of the algorithms can significantly benefit from the understanding of the value of information and its variations across the explored area. Another observation is that while our experiments support the strength of some algorithms, such as Gaussian Processes, they also expose problems such as that their canonical implementations do not scale to the size of problems of interest to current practitioners. We conclude that the proposed benchmark can serve as a catalyst for theoretical research, practical considerations, and efficient implementations in the future. 
\bibliographystyle{abbrv}
\bibliography{references}

\end{document}